# ErfReLU: Adaptive Activation Function for Deep Neural Network

Ashish Rajanand, Pradeep Singh

*Department of Information Technology, National Institute of Technology Raipur, Raipur, 492010, India, arajanand29@gmail.com*
*Department of Computer Science and Engineering, National Institute of Technology Raipur, Raipur, 492010, India, psingh.cs@nitrr.ac.in*

**Abstract**
Recent research has found that the activation function (AF) selected for adding non-linearity into the output can have a big impact on how effectively deep learning networks perform. Developing activation functions that can adapt simultaneously with learning is a need of time. Researchers recently started developing activation functions that can be trained throughout the learning process, known as trainable, or adaptive activation functions (AAF). Research on AAF that enhance the outcomes is still in its early stages. In this paper, a novel activation function 'ErfReLU' has been developed based on the erf function and ReLU. This function exploits the ReLU and the error function (erf) to its advantage. State of art activation functions like Sigmoid, ReLU, Tanh, and their properties have been briefly explained. Adaptive activation functions like Tanhsoft1, Tanhsoft2, Tanhsoft3, TanhLU, SAAF, ErfAct, Pserf, Smish, and Serf have also been described. Lastly, performance analysis of 9 trainable activation functions along with the proposed one namely Tanhsoft1, Tanhsoft2, Tanhsoft3, TanhLU, SAAF, ErfAct, Pserf, Smish, and Serf has been shown by applying these activation functions in MobileNet, VGG16, and ResNet models on CIFAR-10, MNIST, and FMNIST benchmark datasets.

**Keywords**: *Activation function. Deep learning. Image Classification.*

## 1 Introduction

In this modern day of life, deep learning (DL) has emerged as a powerful source to solve different challenging scenarios that occur in the real world, such as image segmentation, object detection, object classification, image restoration, and many more. An artificial neural network (ANN) is a set of neurons. It is comprised of an input layer, a hidden layer, and an output layer. In ANN, the computation of any neuron is divided into two parts. In the first part, the input value is multiplied by a weighted value. Further, in the second part, the decision is made by the activation function based on some threshold value, which defines whether the neurons should fire or not (Apicella et al., 2021). It utilizes these three layers and discovers patterns between input and output with accuracy. To find the pattern, the activation function plays an important role by activating neurons (Gustineli, 2022).

The simplest activation function is the linear AF. An AF plays a very important role in finding a non-linear relationship. Tangent and Sigmoid, ReLU are the most commonly used AF. Some basic properties AF should have are:

### 1.1 Property of AF

Non-linearity: Activation functions play a vital role in dealing with linearity. In the absence of activation function, the data would only move through the network's nodes and layers using linear functions. Irrespective of the number of layers the output is always the result of a linear function reason being the composite of these linear functions also results in a linear function.

Computational cost: Computation of AF should not be complex (Apicella et al., 2021).

Differentiability: Differentiability relates to a function, which is differentiable at every point in a specific domain. A differentiable function can easily calculate the gradient and optimized neuron's weight. A difficult AF will decrease the speed of computation (Maniatopoulos & Mitianoudis, 2021).

Vanishing Gradient: Vanishing gradient problem is one of the major issues, which caused the loss of information during back-propagation. AF should not have a vanishing gradient problem.

Saturation: Saturation is a term that describes the issue where the gradient is being almost zero at some points. This causes difficulty in changing the parameter values(Paul et al., 2022).

Monotonic: There should be no local minima in the graph of a function and after the derivative, the sign of the function should remain unchanged(Biswas et al., 2022).

Less parameter: Most activation functions don't require any additional parameters. It affects the computational calculation of the neuron model(Bingham & Miikkulainen, 2022).

Zero-centred: The zero-centred characteristic facilitates optimization by allowing weight updates to travel in all possible directions (positive and negative), rather than just one. The employment of an activation function identifies whether or not that artificial neuron has been activated(Paul et al., 2022).

In this paper, we discuss several activations and their properties and proposed activation based on the Erf function. Which is used for reducing the dying relu problem (Hao et al., 2020). This paper is divided into several sections. In the first section, we will see different state of art activation functions and in the second section, we will explore some newly arrived AAF.

In the third section, we give a brief introduction to the ErfReLU activation function. We have tested this activation function using deep learning method MobileNet and ResNet18 on CIFER10, MNIST, and FMNIST datasets and their effects.

## 2 Activation Function

Linear Activation Function (LAF): LAF is the simplest function which is defined as $f(x) = kx$, where k is a constant. It is directly proportional to weight or input. It is unable to adapt the non-linearity of data, so the gradient is always constant. Due to this, updating in weight and error is also constant during back-propagation. Therefore, Linear AF is unable to help in adapting complex structure (Kamalov et al., 2021). The researchers started utilizing the logistic (S shape) based activation function as alternative of LAF. It is mainly two types Sigmoid and tanh (hyperbolic tangent function).

### 2.1 Logistic

Sigmoid AF is continuous, bounded between 0 and 1. The gradient of this activation function ranges between 0 and 1. Practically, the gradient becomes zero for the complex neural network. Therefore, error data does not flow via neurons in back-propagation and decreases network performance. However, the solution to this issue is to use better initialization of both the pre-training procedure and the weights

settings(Alkhouly et al., 2021). For example, see Eq. 1 and Fig. 1.

$$sigmoid = \frac{1}{(1+e^{-x})} \quad f'(x) = \frac{1}{(1+e^{-x})^2} \quad (1)$$

$$tanh(x) = \frac{(e^{2x}-1)}{(e^{2x}+1)} \quad f'(x) = 1 - tanh^2(x) \quad (2)$$

The hyperbolic tangent activation function is a scaled version of the sigmoid function. It also has the same S shape (Fig. 2). It is continuous between -1 and +1. This also suffers from a gradient vanishing problem. In contrast to tanh(x), a neural network with arctanh(x) has more severe gradient changes, allowing it to converge more quickly during network training(Sivri et al., 2022). Tanh is defined as E.q.2.

For high positive and small negative values, the Tanh activation function will reach saturation and stop responding to modest changes in the input data. Therefore, the weight won't update and the gradient will simply fade, making DNN unable to finish its training (Lau & Lim, 2019).

## 2.2 ReLU

The ReLU (Maniatopoulos & Mitianoudis, 2021), a quicker AF for learning the complex NN, has emerged as the most useful and popular feature. In comparison to the Sigmoid and Tanh AF, it offers greater deep learning performance. The ReLU retains the characteristics of linear models that make gradient-descent methods an easy way to optimize them because it is a nearly complete representation of the linear function. The ReLU AF executes a threshold operation for each input element, where the value is zero for negative arguments and the variable itself for positive arguments. The Equation for ReLU is given as Eq. 3.

$$ReLU = max(0,x) \quad f'(x) = max(0,1) \quad (3)$$

Due to its simplicity and reduced training time, the Rectified Linear Unit (ReLU). The graphical representation of ReLU is depicted in Fig.3. The ReLU function's drawback is that it has shows the gradient vanishing problem for the negative value of the variable. It also has a dying-relu problem, or non-differentiability at zero, which emerges when a significant negative bias is learned and causes the neuron's output to always be zero regardless of the input (Hao et al., 2020).

## 2.3 Leaky ReLU

LReLU was one of the first rectified-based activation functions built on ReLU. The LReLU function was an attempt to solve the aforesaid ReLU's possible issues. Leaky rectifier activation function enables the small unit of negative value to produce a modest gradient. However, LReLU operates almost equally similar to normal rectifiers. It has a minor effect on network performance.

$$LReLU = \begin{cases} x & if\ x > 0 \\ \alpha x & if\ x \leq 0 \end{cases} \quad f'(x) = \begin{cases} 1 & if\ x > 0 \\ \alpha & if\ x \leq 0 \end{cases} \quad (4)$$

It cannot be utilized for complicated Classification because of its linearity. For certain of the use cases, it performs worse than Sigmoid and Tanh(Maniatopoulos & Mitianoudis, 2021).

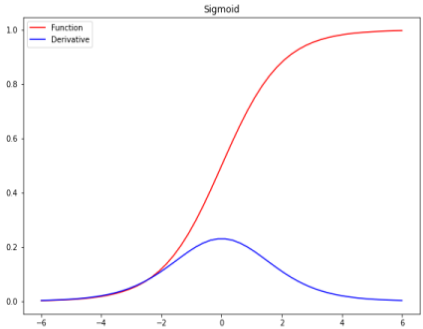
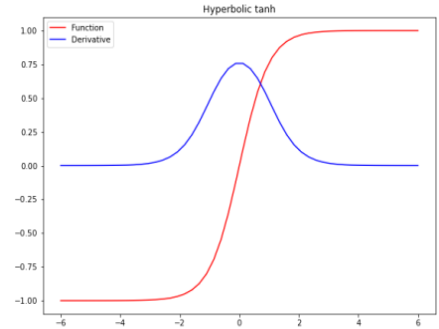
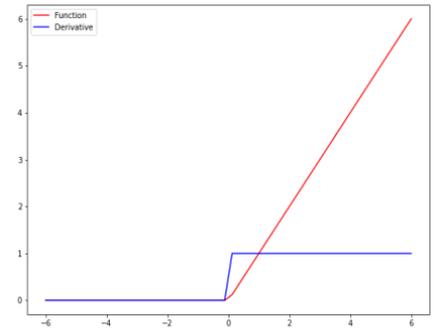

**Fig.1** Sigmoid AF   **Fig.2** Tanh AF   **Fig.3** ReLU

## 2.4 Exponential linear unit (ELU)

The ELU, which was first described by the Clevert et al.(Clevert et al., 2015), is an activation function that preserves identity for inputs that are positive but takes non-zero values for arguments that are negative. It is given as Eq. 5. Where α is a hyper-parameter, which defined the value for negative inputs(Kiliçarslan & Celik, 2021).

$$ELU = \begin{cases} x & if\ x \geq 0 \\ \alpha(e^x - 1) & if\ x < 0 \end{cases} \quad f'(x) = \begin{cases} 1 & if\ x \geq 0 \\ \alpha e^x & if\ x < 0 \end{cases} \quad (5)$$

## 2.5 SILU

The output of the sigmoid function is multiplied with its input in the sigmoid-weighted linear unit (SiLU)(Elfwing et al., 2018) AF as in the output range of $(-0.5, \infty)$. It is defined as Eq.6. For large input, SiLU works as ReLU AF.

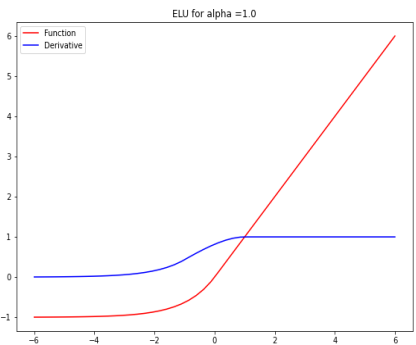
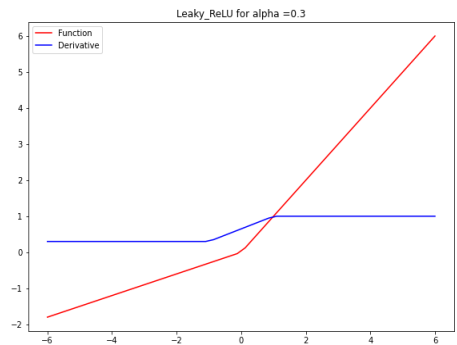
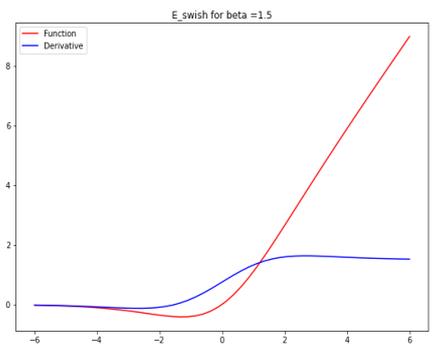

**Fig.4** ELU   **Fig.5** Leaky ReLU   **Fig.6** E_swish

$$SiLU(x) = x \times Sigmoid(x) \quad f'(x) = \frac{1 + e^{-x}(x+1)}{(1+e^{-x})^2} \quad (6)$$

## 2.6 PATS

Zheng et al(Zheng & Wang, 2020) present an AF using inverse hyperbolic tangent similar to Mish, called PATS. It is continuous non-monotonic. They used a parameter to fix the slope in the positive region by updating the parameter. It makes the deep networks more flexible and significantly lowers the risk of overfitting.

$$f(x) = x\arctan(k\pi s(x)) \qquad (7)$$

Where variables s(x) is the sigmoid activation function and k is a constant that brings out from the uniform distribution of the range.

### 2.7 Swish

It is the combination of the input function with sigmoid AF. Some of the characteristics of the Swish function are smoothness, non-monotonicity, and bounding below and unbounded above (Dasgupta et al., 2021).

$$Swish = sigmoid(x) * x = \frac{X}{(1+e^{(-\beta x)})} \qquad (8)$$

$$f'(x) = \frac{1 + e^{(-\beta x)}(x+1)}{(1+e^{(-\beta x)})^2} \qquad (9)$$

It is given as Eq. 8 and also written like Swish(x) = x σ(x). As the value of β increases toward infinity, it behaves like ReLU AF(Ramachandran et al., 2017).

### 2.8 E-swish

A SiLU variant is presented by multiplying the SiLU function by a multiplicative coefficient β. The β parameter needs to tune because E-Swish doesn't have any trainable parameters. The parameter is determined through a search strategy(Alcaide, 2018). E-swish is defined by Eq.10.

$$E - swish(x) = \alpha x sig(x) \quad f'(x) = a\frac{1 + e^{(-\beta x)}(x+1)}{(1+e^{(-\beta x)})^2} \qquad (10)$$

Where hyperparameter β (1<=β<=2) is pre-initialize or learnable. It is also unbounded like ReLU.

### 2.9 Linearly Scaled Hyperbolic Tangent (LiSHT)

Linearly Scaled Hyperbolic Tangent (LiSHT) is combined by multiplying the attribute of ReLU and Tanh activation(Roy et al., 2018) . Non-monotonicity and not being bound are properties of the LiSHT activation function. It is used to tackle the dying ReLU phenomena.

$$LiSHT(x) = x\tanh(x) \quad f'(x) = x(1 - \tanh^2(x)) + \tanh(x) \qquad (11)$$

### 2.10 Mish

Similar to swish Mish AF is also non-monotonic. Identity function, Tanh, and SoftPlus are compounds to make Mish AF. It is based on self-gate property. It resolved the dying ReLU problem by preserving negative value. It doesn't saturate near non-zero values. It provides non-singularity in weight updating due to being continuously differentiable(Misra, 2019).

$$Mish = x\tanh(softplus(x)) \qquad (12)$$

$$f'(x) = \tanh(\ln(1+e^x)) + sech^2(\ln(1+e^x))\frac{x}{1+e^{-x}} \qquad (13)$$

## 3 Adaptive activation functions (AAF)

An adaptive activation function integrated using all the AF, whose shapes are acquired during training based on dataset complexity. A classic AF is converted in the adaptive activation function by changing or by adding simple constraints on the network parameters, which update weights. Kiseľák et al.(Kiseľák et al., 2021) are described a parameterized version of a fixed standard function whose parameter values are learned from data. All adaptive activation functions include a set of trainable parameters that allows AF to adjust shape extremely close to a precisely known function such as ReLU, Sigmoid, and Tanh. In other words, a neural network architecture equipped with AAF behaves similarly to activation functions. For instance, partial derivatives of these new parameters are required to train all the network parameters, including the trainable activation functions. A gradient descent method based on back-propagation is used to update the parameters of AAF. The automatic discovery of parametric activation functions by the PANGAEA, is a genetic parameter tuning method (Bingham & Miikkulainen, 2022).

### 3.1 TanhSoft

It is a zero-centric, non-monotonic, continuous, and trainable activation function. Biswas et al. have proposed three different tanhsoft AAF based on hyper-parameter (Biswas et al., 2021). These are defined as

$$tanhSoft1 = \tanh(\alpha x) * Softplus(x) \qquad (14)$$

$$f'(x) = \tanh(\alpha x)\left[\alpha\cot(\alpha x)\ln(1+e^x) + \frac{1}{1+e^{-x}}\right] \qquad (15)$$

$$tanhSoft2 = x\tanh(\beta e^{\gamma x}) \qquad (16)$$

$$f'(x) = x sec^2 h(\alpha e^{(\beta x)})\alpha e^{(\beta x)} + \tanh(\beta e^{\gamma x}) \qquad (17)$$

$$tanhSoft3 = \ln(1 + e^x \tanh(\delta x)) \qquad (18)$$

$$f'(x) = \frac{1 + \delta \coth(\delta x)}{1 + \coth(\delta x)e^{-x}} \qquad (19)$$

Here, α, β, γ, and δ is hyper-parameter. It is used for controlling the slope of the curve on both the positive and negative axis. TanhSoft with seven baseline activation functions Relu, Leaky ReLU, ELU, Swish, softplus, GeLU, and Mish. TanhSoft outperforms all these activations in different application areas such as object detection, image classification, Machine Translation, and Semantic segmentation.

### 3.2 Sinu-Sigmoidal Linear Unit (SinLU)

Paul et al. proposed a SiLU-based activation function that removes the wiggle by multiplying a sinusoidal function is (x + sinx). SinLU (Paul et al., 2022) AAF is given in Eq.20.

$$SinLU(x) = (x + \alpha \sin\beta x).\sigma(x) \qquad (20)$$

$$f'(x) = \frac{e^{-x}(x+\alpha \sin\beta x)}{(1+e^{-x})^2} + \frac{1+\beta\alpha\cos\beta x}{1+e^{-x}} \qquad (21)$$

Here, α and β are trainable parameters. The parameter α is used to get different shape with amplitude from the rectifier linear unit family and β defined the frequency of the sin function. It is also monotonic and differentiable. It affects the self-gated property of SinLU, so AF works differently from other activation functions. Value of the hyper-parameter of this AAF obtained by the grid search method. They compared SinLU activation with basic activation functions like ReLU, GELU. SinLU outperforms other activation on the MNIST dataset using single-layer feed-forward networks. This AAF is computational expensive, and it can be trapped in the local minimum, making learning extremely difficult(Alkhouly et al., 2021).

### 3.3 TanhLU

New alternatives of Tanh is proposed by integrating Tanh into a linear unit, called TanhLU(Shen et al., 2022), which is motivated by the scaled tanh characteristic and the positive unboundedness of the ReLU. TanhLU is defined as Eq.21.

$$f(x) = \alpha\tanh(\gamma x) + \beta x \quad f'(x) = \alpha\gamma sech^2(\gamma x) + \beta \qquad (21)$$

Here α, β, and γ are the trainable parameters. Shen et al. showed that enlarging the value of the parameter will give a better-weighted gradient. However, if value of β is less than 0.5, it convergences quickly.

### 3.4 SAAF (Shape Auto-tuning Activation Function)

It is non-monotonic and bounded AF. It can gather negative data to enhance generalization ability which prevents and corrects the output distribution from being scattered in the non-negative region. SAAF (Zhou et al., 2021) is given as

$$SAAF = \frac{x}{\frac{x}{\alpha} + e^{\frac{-x}{\beta}}} \quad 0 < \frac{\beta}{\alpha} < 1 \tag{22}$$

$$f'(x) = \frac{\alpha^2(x + \beta)\, e^{\frac{x}{\beta}}}{\beta \left(xe^{\frac{-x}{\beta}} + \alpha\right)^2} \tag{23}$$

It is less saturated compared to the sigmoid and ReLU AF. A saturated region is one with a modest derivative that reduces variation and information propagation to the next layer. Activations are considered normalized if their mean and variance across samples are within prescribed intervals. SAAF provides zero-means property for convergence and decreases the gradient vanishing issue. It can adaptively adjust the output to a normalized distribution using two trainable parameters α and β.

### 3.5 Serf

Serf (Nag & Bhattacharyya, 2021) is inspired by the self-gating property of Swish and Mish, which multiply the output of a non-linear function of an input with the same non-modulated input. Self-gating has an advantage over traditional gating because it only needs a single two-scalar input as opposed to many two-scalar inputs. Serf has properties such as being boundless above, bound beneath, smooth, non-monotonic, and differentiable at every point. Additionally, a limited domain of negative weights are preserved. The serf being unbounded made an effort to get around the saturation issue. Serf mitigates the aforementioned issue in addition to enhancing expressivity and gradient flow by retaining a small portion of negative information. It is a continuous differential AAF.

$$Serf = x\,erf\big(\ln(1 + e^x)\big) \tag{24}$$

$$f'(x) = erf\big(\ln(1 + e^x)\big) + \frac{2xe^{x-\ln^2(e^x+1)}}{\sqrt{\pi}(e^x + 1)} \tag{25}$$

### 3.6 ErfAct and Pserf

ErfAct and Pserf (Biswas et al., 2022) are non-monotonic, continuous, and trainable AF, based on the Gaussian error function. Biswas et al. presented two AAF, by combining the erf function and parametric softplus. It is approximately smooth as ReLU, and outperforms the ReLU AF. Here, Pserf is a parametric version of the Serf Activation function. ErfAct and Pserf are represented by Eq. 26 and Eq. 28 respectively.

$$ErfAct = x\,erf(\alpha e^{\beta x}) \tag{26}$$

$$f'(x) = erf(\alpha e^{\beta x}) + \frac{2\beta \alpha x e^{\beta x - \alpha^2 e^{2\beta x}}}{\sqrt{\pi}} \tag{27}$$

$$Pserf = x\,erf\big(\gamma \ln(1 + e^{\delta x})\big) \tag{28}$$

$$f'(x) = erf\big(\gamma \ln(1 + e^{\delta x})\big) + \frac{2\gamma x e^{\delta - \gamma^3 \ln^2(1+e^{\delta x})}}{\sqrt{\pi}(1 + e^{\delta x})} \tag{29}$$

### 3.7 Smish

An adaptive activation function called Smish inherits the non-monotonic properties of the logish function. Sigmoid(x) is used to reduce the range of values. The logarithm operation is applied to obtain a smooth curve and a flat trend. Smish then multiplies its Tanh operation by x. It has no upper bound and ranges from (-0.25 to ∞). For the larger negative values, it works like the ReLU activation function(Wang et al., 2022).

$$f(x) = x \cdot P(x) \tag{30}$$

$$f'(x) = P(x) + \frac{xe^x \operatorname{sech}^2\big(\ln(1 + sigmoid(x))\big)}{2e^{2x} + 3e^x + 1} \tag{31}$$

$$\text{where} \quad P(x) = \tanh\big[\ln(1 + sigmoid(x))\big] \tag{32}$$

### 3.8 IpLU

IpLU (Wu et al., 2021) is a parametric version of the softsign function. Unlike Tanh, softsign has polynomial convergence, whereas the Tanh function has exponential convergence. It is used in the negative region and a parameter α is used to get smaller slop. The Linear function is used to avoid exploding gradient problems. It is zero centric, monotonic and its derivative is also zero centric.

$$f(x) = \begin{cases} x & if\, x \geq 0 \\ \dfrac{x}{1 + |x|^\alpha} & if\, x < 0 \end{cases} \tag{33}$$

$$f'(x) = \begin{cases} 1 & if\, x \geq 0 \\ \dfrac{1 + |x|^\alpha + x^2 \alpha |x|^{(\alpha - 2)}}{(1 + |x|^\alpha)^2} & if\, x < 0 \end{cases} \tag{34}$$

## 4 Proposed Activation function

After extensive literature survey, we obtained that most of the AAFs are reducing the effect of Dying ReLU phenomena at negative values(Zhou et al., 2021)(Shen et al., 2022)(Nag & Bhattacharyya, 2021) (Wang et al., 2022). ReLU activation, in which the negative part is not being considered. We came up with an activation function, which not only solved the dying ReLU by applying the function at the negative side (x<0) but also removes the negative saturation of sigmoid. It is achieved by piecewise combination of Erf and ReLU functions, called ErfReLU. The use of the erf function is inspired by the ErfAct activation function Eq.19. Proposed ErfReLU is a monotonic, zero-centered activation function. Erf is a trainable AF which is an approximation of ReLU with having negative size of ReLU. Biswas et al. (Biswas et al., 2022) have used erf & Pserf with α,β,γ, and δ as their parameter that controls the slop of the curve. The proposed AAF only uses one parameter. The Erf function is the gauss error function. It is bound below, which is represented in Eq.35.

$$erf(x) = \frac{2}{\sqrt{\pi}} \int_0^x e^{-t^2}\, dt \tag{35}$$

The derivative of the above function is given as

$$erf'(x) = \frac{2}{\sqrt{\pi}} e^{-x^2} \tag{36}$$

ErfReLU is defined as

$$f(x) = \begin{cases} x & if\ x \geq 0 \\ \alpha\, \operatorname{erf}(x) & if\ x < 0 \end{cases} \tag{37}$$

The derivative of ErfReLU is defined as Eq.23.

$$f'(x) = \begin{cases} 1 & if\ x \geq 0 \\ \dfrac{\alpha 2 e^{-x^2}}{\sqrt{\pi}} & if\ x < 0 \end{cases} \quad (38)$$

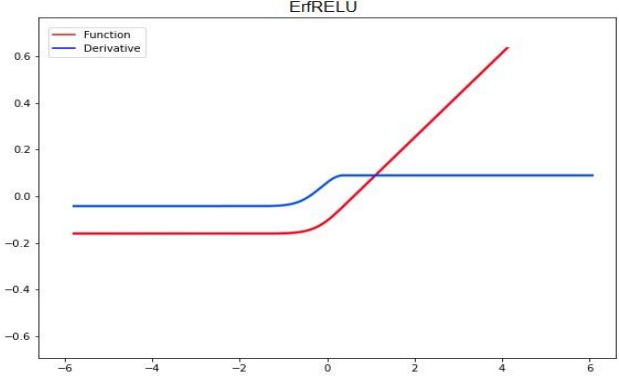

**Fig.10** Proposed ErfReLU activation function and its derivative.

Here α is a parameter, used to control the slope of the function. The graph of ErfReLU and its derivative is shown in fig.10. It is continuous and above unbounded but bounded below AAF. The positive region of the ErfReLU function works as the ReLU activation function. It preserves a small amount of negative information to prevent the occurrence of the Dying ReLU phenomena and achieves the better accuracy with only one trainable parameter.

### 4.1 Properties of the Proposed Function

- Smoothness: The Gaussian error function and ReLU are differentiable everywhere, so ErfReLU is also differentiable, which makes it a smooth function.
- Non-linearity: The Gaussian error function is a non-linear function, which introduces non-linearities into the network that helps to learn more complex functions and relationships between the input and output.
- Saturation: The Gaussian error function can become saturated for large input values, meaning that the output of the function approaches a constant value as the input becomes very large. However, ReLU is not saturated for large input(Nag & Bhattacharyya, 2021). This exploits the benefits of both Erf and ReLU to proposed ErfReLU to learn complex functions
- Boundedness: ErfReLU is lower bound as it is using the erf function. There is upper boundedness in the ErfReLU, due to owing the properties of the ReLU AF(Shen et al., 2022).

## 5 Experiments and Results

In this section, we describe the experiments performed on the above-defined adaptive activation functions on diverse datasets using different deep learning models. We compared our suggested activations to 9 prominent trainable and state of art activation functions on deep learning application areas such as image classification and machine translation. The experimental data shows that ErfReLU has significant improvement over other AAF and traditional activations in the deep learning problem. We examine the performance by simply switching the activation functions while maintaining all other network and training parameter constants.

In the adaptive activation function, the value of the parameter is depending upon input dataset, it used for updating the activation function and its shape. MobileNetV1, ResNet18, VGG16 and machine translation seq2seq model are used in performance analysis for AAFs. The experiments are performed using the PyTorch framework over a desktop system (Nvidia Quardo Rtx 5000 GPU had 128 GB RAM) consisting of 16 GB GPU.

### 5.1 Image Classification

We first evaluate our proposed activation functions on diverse image classification dataset such as the MNIST, FMNIST, and CIFER10. We conduct comparative studies on two types of representative architectures, a large model (like VGG and ResNet) and a lightweight model (MobileNet), maintaining all parameter constants. MobileNet V1 is a lightweight version of the traditional convolutional layers called "depthwise separable convolution" to reduces the number of parameters, memory consumption, and computation.

The ResNet 18 architecture consists of 18 layers, including convolutional layers, batch normalization layers, and rectified linear unit (ReLU) activation layers. The residual connections help to preserve the gradients flowing through the network, allowing the network to learn effectively even when it has many layers. The architecture of VGG16 is composed of a series of convolutional layers, followed by max-pooling layers, and several fully connected layers. We used the Cross Entropy Loss function for error calculation and classification of images. Adam optimizer is used with a learning rate of 0.001. We train the model for 100 epochs with batch size 128. Transformation is applied on the image with the size of 32 and 4-pixel zero-padding.

#### 5.1.1 CIFAR-10

CIFAR-10 is a dataset of size 60,000 32x32 colour images in 10 classes, with 6,000 images per class. There are 50,000 training images and 10,000 test images. The classes are airplane, automobile, bird, cat, deer, dog, frog, horse, ship, and truck. Table 1 provides the classification accuracy of different models on the defined adaptive activation function. Our proposed activation function shows significant improvement over other adaptive activation based on several observations on MobileNet and ResNet. ErfReLU trounces all other AAFs. It achieves highest accuracy of 92.78% and 94.04% over Mobinenet and Resnet models, respectively. Only on VGG16 model, ErfAct achieved greater accuracy than ErfReLU. Table 2 provides the categorization outcomes for CIFAR10 as determined by the activation functions ReLU, Swish, Mish, LiSHT and ErfReLU. The ErfReLU outperforms the other defined the state-of-the-art activation function over Mobinenet and Resnet models. It also shows significant improvement over Swish, LiSHT, activation functions on VGG16 model, however ReLU has better performance.

**Table 1** Comparison of testing accuracy of different adaptive activation functions on the CIFER10 dataset.

| S/N. | Activation function | Parameter | MobileNet | ResNet | VGG 16 |
|---|---|---|---|---|---|
| 1 | Tanhsoft 1 | α: 0.87 | 92.24 | 93.67 | 92.58 |
| 2 | Tanhsoft 2 | β: 0.75, γ: 0.75 | 92.39 | 93.44 | 92.87 |
| 3 | Tanhsoft 3 | α: 0.85 | 92.45 | 93.43 | 92.57 |
| 4 | TanhLU | α: 1, β: 0.5, γ=2 | 90.86 | 90.35 | 89.84 |
| 5 | SAAF | α: 3, β: 2 | 91.64 | 90.58 | 86.28 |
| 6 | ErfAct | α: 0.75, β: 0.75 | 92.52 | 93.82 | 93.24 |
| 7 | Pserf | α: 1.25, β: 0.85 | 92.67 | 93.84 | 92.76 |

| 8 | Smish | α :0.95, β: 1.2 | 92.58 | 93.83 | 92.46 |
| 9 | Serf | α: 1.25, β: 0.80, γ=2 | 92.67 | 93.63 | 92.73 |
| 10 | ErfReLU | α: 0.882267 | **92.78** | **94.04** | **93** |

**Table 2** Comparison of testing accuracy of our proposed AAF with different state of art activation functions over CIFER10 dataset.

| CIFER10 | Activation function | MobileNet | ResNet | VGG16 |
|---|---|---|---|---|
| 1. | ReLU | 90.13 | 93.63 | 93.9 |
| 2. | Swish | 90.84 | 93.64 | 92.8 |
| 3. | Mish | 90.78 | 93.43 | 93.1 |
| 4. | LiSHT | 85.92 | 90.15 | 86.5 |
| 5. | **ErfReLU** | **92.78** | **94.04** | **93** |

### 5.1.2 MNIST

The MNIST (Biswas et al., 2022) dataset of handwritten digits contains 70K images (60k for training and 10k for testing purposes), and each image size is a 32*32-pixel grayscale image. Table 3 shows the comparative result of AAF with proposed. It can be seen in table 4, the ErfReLU shows significant improvement over the state-of-the-art activation function.

**Table 3** Comparison of testing accuracy of different AAF on the MNIST dataset.

| MNIST | Activation function | Parameter | MobileNet | ResNet | VGG16 |
|---|---|---|---|---|---|
| 1. | Tanhsoft1 | α: 0.87 | 98.74 | 99.15 | 99.5 |
| 2. | Tanhsoft2 | β: 0.75, γ: 0.75 | 98.6 | 99.09 | 99.6 |
| 3. | Tanhsoft3 | α: 0.85 | 98.59 | 99.12 | 99.6 |
| 4. | TanhLU | α: 1, β: 0.5, γ=2 | 94.64 | 99.35 | 99.4 |
| 5. | SAAF | α: 3, β: 2 | 89.04 | 98.28 | 99.4 |
| 6. | ErfAct | α: 0.75, β: 0.75 | 98.85 | 99.12 | 99.4 |
| 7. | Pserf | α: 1.25, β: 0.85 | 98.55 | 99.12 | 99.6 |
| 8. | Smish | α :0.95, β: 1.2 | 98.59 | 99.12 | 99.3 |
| 9. | Serf | α: 1.25, β: 0.80, γ=2 | 98.47 | 99.17 | 99.5 |
| 10. | **ErfReLU** | α: 0.882267 | **98.85** | **98.99** | **99.6** |

**Table 4** Comparison of testing accuracy of different state of art activation functions on the MNIST dataset.

| MNIST | Activation function | MobileNet | ResNet | VGG16 |
|---|---|---|---|---|
| 1. | ReLU | 98.92 | 99.01 | 99.4 |
| 2. | Swish | 99.05 | 99.09 | 99.6 |
| 3. | Mish | 98.73 | 99.13 | 99.5 |
| 4. | LiSHT | 98.80 | 99.13 | 99.6 |
| 5. | **ErfReLU** | **98.85** | **98.99** | **99.6** |

### 5.1.3 FMNIST

The FMNIST(Biswas et al., 2022) dataset consists of 70,000 handwritten digit images, with 60,000 images allocated for training and 10,000 for testing. Each image is a 32*32-pixel grayscale image. As presented in Table 5, the ErfReLU activation function outperforms the other activation functions when used with Mobilenet and Resnet models. However, in VGG16, the Smish adaptive activation function displays better performance compared to other adaptive activation functions. Moreover, Table 6 demonstrates that the ErfReLU activation function demonstrates a significant improvement over the state-of-the-art activation function.

**Table 5** Comparison of testing accuracy of different adaptive activation functions on the FMNIST dataset.

| S/N. | Activation function | Parameter | Mobile-Net | ResNet | VGG16 |
|---|---|---|---|---|---|
| 1. | Tanhsoft1 | α: 0.87 | 92.24 | 93.71 | 93.3 |
| 2. | Tanhsoft2 | β: 0.75, γ: 0.75 | 92.39 | 92.44 | 93.6 |
| 3. | Tanhsoft3 | α: 0.85 | 92.45 | 93.58 | 93.6 |
| 4. | TanhLU | α: 1, β: 0.5, γ=2 | 90.86 | 92.9 | 92.9 |
| 5. | SAAF | α: 3, β: 2 | 91.64 | 92.97 | 98.6 |
| 6. | ErfAct | α: 0.75, β: 0.75 | 92.52 | 93.65 | 93.2 |
| 7. | Pserf | α: 1.25, β: 0.85 | 92.67 | 93.66 | 93.4 |
| 8. | Smish | α :0.95, β: 1.2 | 92.58 | 93.67 | 93.8 |
| 9. | Serf | α: 1.25, β: 0.80, γ=2 | 92.67 | 93.46 | 93.7 |
| 10. | **ErfRELU** | α: 0.882267 | **92.78** | **93.72** | **93.3** |

**Table 6** The testing accuracy of various state-of-the-art activation functions are compared on the FMNIST dataset.

| FMNIST | Activation function | MobileNet | ResNet | VGG16 |
|---|---|---|---|---|
| 1. | ReLU | 92.59 | 93.73 | 93.7 |
| 2. | Swish | 92.74 | 93.42 | 92.8 |
| 3. | Mish | 92.55 | 93.56 | 93.1 |
| 4. | LiSHT | 90.59 | 92.38 | 93.2 |
| 5. | **ErfReLU** | **92.78** | **93.72** | **93.3** |

### 5.2 Machine Translation

Machine translation is a type of technology that automatically translates text from one language to another. It is used in a variety of application areas such as website localization, document translation, and chatbots. The validation of AAF's effectiveness in machine translation has been carried out through English-to-German translation experiments, utilizing the standard Multi30k dataset for training and testing. This dataset comprises 29k sentences for training, as well as 1k each for validation and testing purposes. A Long Short-Term Memory (LSTM) auto-encoder network is utilized as the benchmark Seq2Seq model for English-to-German translation. The encoder and decoder have an embedding size of 300, and both are equipped with a dropout factor of 0.5. The language translation model was trained for 25 epochs using a learning rate of 0.001 and a batch size of 256, while utilizing the Adam optimizer for cross-entropy loss training(Dubey et al., 2022).

The network's performance on the dataset was evaluated using the BiLingual Evaluation Understudy (BLEU) score measure. Table 7 illustrates the BLEU score achieved by the adaptive activation function. When compared to other adaptive activation functions, the ErfReLU activation function displayed a notable improvement in performance. Nevertheless, the SAAF activation function exhibited superior performance compared to other activation functions.

**Table 7.** Comparison of BLEU score of different adaptive activation functions over the Multi30k dataset.

| S.N. | Activation function | Parameter | BLEU score |
|---|---|---|---|
| 1. | Tanhsoft1 | $\alpha$: 0.87 | 20.82 |
| 2. | Tanhsoft2 | $\beta$: 0.75, $\gamma$: 0.75 | 21.27 |
| 3. | Tanhsoft3 | $\alpha$: 0.85 | 20.67 |
| 4. | TanhLU | $\alpha$: 1, $\beta$: 0.5, $\gamma$=2 | 19.98 |
| 5. | SAAF | $\alpha$: 3, $\beta$: 2 | 13.05 |
| 6. | ErfAct | $\alpha$: 0.75, $\beta$: 0.75 | 20.81 |
| 7. | Pserf | $\alpha$: 1.25, $\beta$: 0.85 | 21.23 |
| 8. | Smish | $\alpha$: 0.95, $\beta$: 1.2 | 21.43 |
| 9. | Serf | $\alpha$: 1.25, $\beta$: 0.80, $\gamma$=2 | 19.94 |
| 10. | ErfRELU | $\alpha$: 0.882267 | 19.54 |

## 6 Conclusion

In this paper, adaptive activation functions that have been explored, the function that focuses on the ones that can be trained, are studied. These activation function attempts to avoid issues that arise during the training and optimization processes, such as learning saturation, exploding vanishing gradients, and dying problems. The control parameters are used in adaptive AFs to alter the efficiency of nonlinear transformations. We proposed an activation function, called ErfReLU, which combines the attribute of ReLU and Erf gives better convergence than state of art AF such as ReLU and other adaptive activation functions. An experimental analysis of nine adaptive AFs is performed, which can be used in many different areas of Deep learning. The CIFAR10, MNIST, and FMNIST dataset is used to evaluate the effectiveness of the adaptive AFs with MobileNet, ResNet, and Vgg16. Machine translation is performed on a multi3k dataset on an LSTM-based autoencoder. In general, the adaptive Afs show better convergence as they can adapt the data faster by learning the parameter from the data. ErfReLU has shown better convergence in all nine adaptive activation functions, although in some cases, Pserf showed better convergence.

**Declaration of Competing Interest**

The authors declare that they have no known competing financial interests or personal relationships that could have appeared to influence the work reported in this paper.